\documentclass{article}
\usepackage{spconf,amsmath,graphicx}

\usepackage{subfigure} 
\graphicspath{{./image/}}

\usepackage{amssymb}
\DeclareMathOperator*{\relu}{ReLU}

\DeclareMathOperator*{\sigmoid}{sigmoid}
\DeclareMathOperator*{\intersection}{intersection}
\DeclareMathOperator*{\union}{union}
\usepackage{epstopdf}
\usepackage{tabularx}
\usepackage{multirow}
\usepackage{textcomp}
\usepackage{url}
\usepackage{lipsum}
\usepackage{makecell}

\usepackage{bm}
\usepackage{calc}

\usepackage{color}
\usepackage{dsfont}

\title{Unifying Isolated and Overlapping Audio Event Detection with Multi-Label Multi-Task Convolutional Recurrent Neural Networks}
\name{\begin{tabular}{c}Huy Phan$^{\ast\ddagger}$\thanks{The research was performed when H. Phan was at the University of Oxford and supported by the NIHR Oxford Biomedical Research Centre. Corresponding author: \tt h.phan@kent.ac.uk}, Oliver Y. Ch\'{e}n$^{\ast}$, Philipp Koch$^{\dagger}$, Lam Pham$^{\ddagger}$ \\
		Ian McLoughlin$^{\ddagger}$, Alfred Mertins$^{\dagger}$, and Maarten De Vos$^{\ast}$ \end{tabular}}
\address{$^\ddagger$ University of Kent, School of Computing, UK \\
	$^\ast$ University of Oxford, Department of Engineering Science, UK \\
	$^\dagger$ University of L\"ubeck, Institute for Signal Processing, Germany \\
}

\begin{document}
\ninept
\maketitle
\begin{abstract}
We propose a multi-label multi-task framework based on a convolutional recurrent neural network to unify detection of isolated and overlapping audio events. The framework leverages the power of convolutional recurrent neural network architectures; convolutional layers learn effective features over which higher recurrent layers perform sequential modelling. Furthermore, the output layer is designed to handle arbitrary degrees of event overlap. At each time step in the recurrent output sequence, an output triple is dedicated to \emph{each} event category of interest to jointly model event occurrence and temporal boundaries. That is, the network jointly determines whether an event of this category occurs, and when it occurs, by estimating onset and offset positions at each recurrent time step. We then introduce three sequential losses for network training: multi-label classification loss, distance estimation loss, and confidence loss. We demonstrate good generalization  on two datasets: ITC-Irst for isolated audio event detection, and TUT-SED-Synthetic-2016 for overlapping audio event detection.
\end{abstract}
\begin{keywords}
Audio event detection, isolated sound, overlapping sound, multi-label, multi-task, convolutional recurrent neural network
\end{keywords}
\vspace{-0.15cm}
\section{Introduction}
\label{sec:intro}
\vspace{-0.2cm}
Audio event detection (AED) \cite{Stowell2015,Mesaros2018} is an important research area within the wider machine hearing field \cite{Wang2006, Lyon2010}, aiming at determining when and which target events occur in continuous audio. This task has recently attracted significant attention in the research community, manifested by rapidly increasing numbers of participants in related international challenges over the past few years \cite{dcaseweb}. 
Ideally, event instances of different categories of interest would occur in isolation so that there is at most one such occurrence at any time point in the signal~\cite{Phan2017e,McLoughlin2017}. However, in practice, they may occur at the same time, leading to partial or full temporal overlap~\cite{Cakir2017,Parascandolo2016}, sometimes called polyphonic AED. Due to the mixture of multiple acoustic sources, detection of overlapping events is much more challenging than detection of isolated ones.

Isolated and overlapping AED literature often appear to derive from two separate methodological streams. For the former, there is a large body of work covering different perspectives: noise robustness \cite{McLoughlin2015,McLoughlin2017,Dennis2013a}, multichannel and multimodal fusion \cite{Kuerby2016,Imoto2017,Phan2015a}, weak labelling \cite{Kumar2017, Kong2018}, early event detection \cite{phan2015b, Phan2018a, McLoughlin2018}, event detection under scarcity scenarios \cite{Phan2018b,Lim2017}, as well as false positive reduction \cite{phan2017d, McLoughlin2017}. Particularly, 
the multitasking approach that jointly performs event detection and event boundary estimation \cite{Phan2015, Phan2017e, Xia2018} has  demonstrated state-of-the-art performance on different benchmark datasets.
In the latter stream, overlapping events are either separated using source separation methods \cite{Heittola2013, Mesaros2015} prior to detection,  or recognized via a selection of local spectral features \cite{Dennis2013a, McLoughlin2015, McLoughlin2017}. The most successful approach appears to be to directly classify the mixtures via multi-label classification \cite{Tran2011,Phan2016e,Cakir2015,Cakir2017,Parascandolo2016}. But both streams have one aspect in common: they are efficient when coupled with underlying deep learning models \cite{Cakir2017, Kao2018, Phan2017e, Phan2018b}, particularly convolutional recurrent neural networks (CRNN) \cite{Cakir2017, Kao2018}. This is partly due to their power in feature learning and partly due to their capability in performing complex modelling tasks, i.e. multi-label and multi-task. However, there exists a methodological gap between them. Audio events intrinsically possess temporal structures, and tailoring a network's output layer and loss functions for structure modelling has been shown to be efficient for the isolated AED \cite{Phan2015, Phan2017e, Xia2018}. However, this capacity has been uncharted for overlapping AED, and it remains questionable how to generalize a network's output layer and tailor its loss functions \cite{Phan2018b,Phan2017e} to accommodate arbitrary event overlap, i.e. from one to the maximum number of simultaneous or partially-simultaneous target events. Bridging this gap would allow us to unify how isolated and overlapping AED is trained and operated.

To this end, we present a multi-label multi-task CRNN framework to homogeneously deal with isolated and overlapping events. The network body makes use of a CRNN architecture as it has been shown to be efficient for both isolated \cite{Kao2018,Cakir2017} and overlapping \cite{Cakir2017} AED. The idea is to use the convolutional layers to learn good time-frequency invariant features over which recurrent layers are leveraged to incorporate a long temporal context, i.e. hundreds of audio frames. The network sequential output layer is designed to accommodate all possible event overlaps. At each time step of the recurrent output, we tailor a set of output triplets each of which is dedicated to one event category of interest. The output consists of three elements: event activity, event onset distance, and event offset distance, to allow the network to determine whether or not an event of this category is happening at the current time index, and estimate the distances to its boundary, i.e. event onset and offset position, at the same time. As one output triplet is tied to one specific category, inference for all target event categories can be carried out individually no matter how many events of different classes occur concurrently. For training, three types of loss are proposed: sequential multi-label classification loss, sequential distance estimation loss, and sequential confidence loss. Combining the three losses, the network is penalized for both mistakes it makes on event activity determination and event boundary estimation, integrated over all time steps of the recurrent output layer. 

\vspace{-0.2cm}
\section{The Multi-Label Multi-Task CRNN}
\label{sec:approach_overview}
\vspace{-0.2cm}

\begin{figure} [!t]
	\centering
	\includegraphics[width=0.95\linewidth]{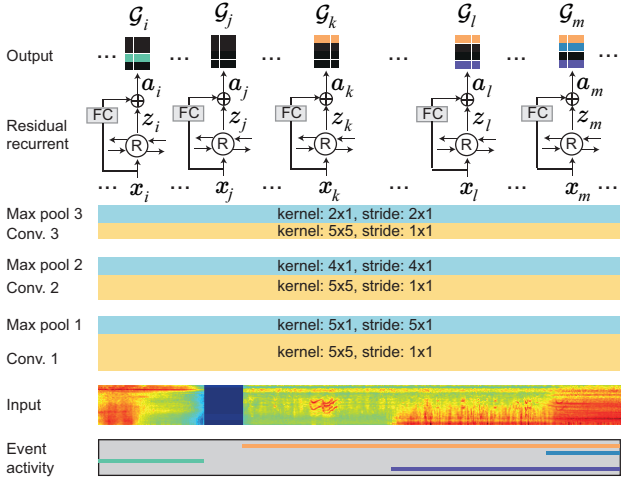}
	\vspace{-0.2cm}
	\caption{Overview of the proposed multi-label multi-task CRNN.}
	\label{fig:overview}
	\vspace{-0.3cm}
\end{figure}

The proposed network architecture is illustrated in Fig. \ref{fig:overview} and is described in detail in the following sections. 
\vspace{-0.2cm}
\subsection{Input}
\vspace{-0.2cm}
An audio signal, sampled at 44100 Hz, is converted into a log Mel-scale spectrogram using $M=40$ Mel-scale filters in the frequency range of [50, 22050] Hz. In addition, a window (i.e. frame) size of 40 ms with 50\% overlap is used. $C$ different event categories are considered in total. Since events from any of these categories may happen at a certain audio frame, in order to accommodate all possible occurrences it is necessary to annotate each audio frame with a set of $C$ triplets $\mathcal{G}=\{(y^c, p^c, q^c)\}$, $1 \le c \le C$, one of which is dedicated for each event category. $y^c \in \{0,1\}$ equals to one if an event of class $c$ is active at the current audio frame and equals to zero otherwise. $p^c,q^c  \in R_{+}$ denote the distances from the current frame to the event onset and offset if it is active and are normalized to [0,1]. $p^c$ and $q^c$ are forced to be zero when the event is absent. 

As a long context is crucial for audio event detection \cite{Cakir2017,Kao2018}, we use an audio segment of $T=512$ frames as an input to the network. Hence, one sample, i.e. one audio segment, is represented by a time-frequency image $\mathbf{S} \in \mathbb{R}^{M \times T}$ and associated with a sequence of $T$ triplet sets $\mathfrak{G} = (\mathcal{G}_1, \mathcal{G}_2, \ldots, \mathcal{G}_T)$ where  $\mathcal{G}_t = \{(y^c_t, p^c_t, q^c_t)\}$, $1 \le c \le C$, as described above, denotes the annotation of the frame at the time index $t$, $1 \le t \le T$. The network therefore plays the role of mapping: $\mathbf{S} \mapsto \mathfrak{G}$, which is multi-label (i.e. multiple classes may be active concurrently) and multi-task (i.e. joint modelling of event activity and event boundary).

\vspace{-0.2cm}
\subsection{Convolutional layers}
\vspace{-0.2cm}
The convolutional block of the network essentially consists of three convolutional layers, each followed by a max pooling layer. The convolutional layers are commonly designed to have $F$ two-dimensional convolutional filters of size $5 \times 5$ with the stride set to one in both temporal and spectral directions during the convolution operation. Zero-padding (also known as \emph{SAME} padding) is used in order to maintain a temporal size equal to $T$. After convolution, batch normalization \cite{Ioffe2015} is applied on the feature maps, followed by Rectified Linear Units (ReLU) activation \cite{Nair2010}.

The three max pooling layers aim to improve spectral invariance while keeping the temporal size unchanged. Their pooling kernel size are set to $5 \times 1$, $4\times 1$, and $2\times 1$ with stride of $5\times 1$, $4\times 1$, $2\times 1$, respectively. With these settings, the spectral dimension is reduced from an input of $M=40$ to $8 \rightarrow 2 \rightarrow 1$ after the pooling layers, respectively. Concatenating all $F$ pooled feature maps after the last pooling layer, we have transformed the original input into a convolutional image feature $\mathbf{X} \in \mathbb{R}^{F \times T}$.

\vspace{-0.2cm}
\subsection{Residual recurrent layer}
\vspace{-0.2cm}
The above convolutional output $\mathbf{X}$ can be interpreted as a sequence of $T$ convolutional feature vectors, i.e. $\mathbf{X}~\equiv~(\mathbf{x}_1,\mathbf{x}_2,\ldots, \mathbf{x}_T)$ where $\mathbf{x}_t~\in~\mathbb{R}^F$, $1 \le t \le T$. A bidirectional recurrent layer is then used to read the sequence of convolutional feature vectors into the sequence of recurrent feature vectors $\mathbf{Z}~\equiv~(\mathbf{z}_1,\mathbf{z}_2,\ldots, \mathbf{z}_T)$, where 
\begin{align}
\mathbf{z}_t &= [\mathbf{h}^{\text{b}}_t \oplus \mathbf{h}^{\text{f}}_t]\mathbf{W}_{z} + \mathbf{b}_{z},
\label{eq:rnn_output} \\
\mathbf{h}^{\text{f}}_t &= \mathcal{H}(\mathbf{x}_t\,, \mathbf{h}^{\text{f}}_{t-1}), \label{eq:rnn_hidden_forward} \\
\mathbf{h}^{\text{b}}_t &= \mathcal{H}(\mathbf{x}_t\,, \mathbf{h}^{\text{b}}_{t+1}).
\label{eq:rnn_hidden_backward}
\end{align}
Here, $\mathbf{h}^{\text{f}}_t, \mathbf{h}^{\text{b}}_t \in \mathbb{R}^H$ represent the forward and backward hidden state vectors of size $H$ at recurrent time step $t$, respectively while $\oplus$ indicates vector concatenation. 
$\mathbf{W}_{z} \in \mathbb{R}^{2H \times 2H}$ denotes a weight matrix and $\mathbf{b}_{z} \in  \mathbb{R}^{2H}$ denotes bias terms. $\mathcal{H}$ represents the hidden layer function of the recurrent layer and we use Gated Recurrent Units (GRU) \cite{Cho2014} here to realize the function $\mathcal{H}$.

As a recurrent output vector $\mathbf{z}_t$ is expected to have context information from the entire sequence, to allow the network to explore possible combinations of local convolutional features $\mathbf{x}_t$ and contextual recurrent features $\mathbf{z}_t$, we aggregate them via a residual connection. As $\mathbf{x}_t \in \mathbb{R}^F$ and $\mathbf{z}_t \in \mathbb{R}^{2H}$ may have different sizes in practice, we transfer $\mathbf{x}_t$ through a fully-connected layer with a weight matrix $\mathbf{W}_{x} \in \mathbf{R}^{F\times 2H}$ and bias term $\mathbf{b}_{x} \in \mathbf{R}^{2H}$ to make their sizes compatible. The final residual feature vector at time step $t$ is
\begin{align}
\mathbf{a}_t = \relu(\mathbf{\mathbf{x}_t W}_{x} + \mathbf{b}_{x})  \oplus \mathbf{z}_t.
\end{align}
Batch normalization \cite{Ioffe2015} is also applied to the fully-connected layer of the residual connection.

\vspace{-0.2cm}
\subsection{Output layer}
\vspace{-0.2cm}
The output layer consists of $T \times C \times 3$ entries in total which are orderly arranged in the output sequence $\hat{\mathfrak{G}} = (\hat{\mathcal{G}}_1, \hat{\mathcal{G}}_2, \ldots, \hat{\mathcal{G}}_T)$. At time index $t$, $\hat{\mathcal{G}}_t = \{(\hat{y}^c_t, \hat{p}^c_t, \hat{q}^c_t)\}$, $1 \le c \le C$, is the set of $C$ output triplets, one dedicated to each event category. $\hat{y}^c_t$ indicates how likely an event of class $c$ is occurring at $t$ while $\hat{p}^c_t$ and $\hat{q}^c_t$ estimate the distances to its onset and offset from $t$. To obtain the output $\hat{\mathcal{G}}_t$ at time index $t$, the residual output $\mathbf{a}_t $ is transferred through a single fully-connected layer with \emph{sigmoid} activation:
\begin{align}
\mathbf{o}_{\hat{\mathcal{G}}_t} = \sigmoid(\mathbf{\mathbf{a}}_t \mathbf{W}_{a} + \mathbf{b}_{a}),
\end{align}
where $\mathbf{W}_{a} \in \mathbb{R}^{2H\times3C}$ and $\mathbf{b}_{a} \in \mathbb{R}^{3C}$.  $\mathbf{o}_{\hat{\mathcal{G}}_t} \in [0,1]^{3C}$ is the flattened vector including the entries of $\hat{\mathcal{G}}_t$ in a pre-determined order.

\vspace{-0.2cm}
\subsection{Losses}
\vspace{-0.2cm}
Similar to \cite{Phan2018b,Phan2017e}, for network training, we want to penalize the network on both tasks: event activity determination and event boundary estimation. 
Assume output sequence $\hat{\mathfrak{G}} = (\hat{\mathcal{G}}_1, \hat{\mathcal{G}}_2, \ldots, \hat{\mathcal{G}}_T)$ is obtained from the network given an input $\mathbf{S}$ associated with groundtruth $\mathfrak{G} = ({\mathcal{G}}_1, {\mathcal{G}}_2, \ldots, {\mathcal{G}}_T)$. Since the event activity determination task is addressed as a multi-label classification problem, i.e. multiple events of different classes may be present at the same time, the multi-class cross-entropy loss cannot be used here. Therefore, we interpret the multi-label classification problem as multiple binary classification problems and employ \emph{binary} cross-entropy loss penalization. Furthermore, integration over the network's output at different time steps is necessary to take into account all possible misclassifications. The \emph{sequential multi-label classification loss} is
\vspace{-0.25cm}
\begin{align}
E_{\text{class}}(\mathfrak{G}, \hat{\mathfrak{G}})\!=\! \frac{1}{T}\!\sum_{t=1}^{T}\!\sum_{c=1}^{C}\!\!\Big(\!\!-\!y^c_t\log(\hat{y}^c_t)\!-\!(1\!-\!y^c_t)\log(1\!-\!\hat{y}^c_t)\!\Big). \label{eq:class_loss}
\end{align}
\vspace{-0.2cm}

Similarly, the \emph{sequential distance loss} induced by errors in event onset and offset distance estimation is given by
\vspace{-0.2cm}
\begin{align}
E_{\text{dist}}(\mathfrak{G}, \hat{\mathfrak{G}})=\frac{1}{T}\sum_{t=1}^{T}\sum_{c=1}^{C}\Big(\|p^c_t - \hat{p}^c_t\|^2_2 + \|q^c_t - \hat{q}^c_t\|^2_2 \Big). \label{eq:dist_loss}
\end{align}
The event boundary estimation errors can also be quantified as the 
$\{{\intersection}:{\union}\}$ ratio of the truth boundary and the estimated boundary \cite{Phan2018b,Phan2017e} and further penalized using the \emph{sequential confidence loss}:
\vspace{-0.1cm}
\begin{align}
E_{\text{conf}}(\mathfrak{G}, \hat{\mathfrak{G}})&\!=\!\frac{1}{T}\!\sum_{t=1}^{T}\!\sum_{c=1}^{C}\!\Big\|y^c_t\!-\!\frac{\intersection(\mathfrak{G}, \hat{\mathfrak{G}})}{\union(\mathfrak{G}, \hat{\mathfrak{G}})}\Big\|^2_2 \nonumber \\
&\!=\!\frac{1}{T}\!\sum_{t=1}^{T}\!\sum_{c=1}^{C}\!\Big\|y^c_t\!-\!\frac{\min(p^c_t, \hat{p}^c_t)\!+\!\min(q^c_t, \hat{q}^c_t)}{\max(p^c_t, \hat{p}^c_t)\!+\!\max(q^c_t, \hat{q}^c_t)}\Big\|^2_2.
 \label{eq:conf_loss} 
\end{align}
Note that we do not use the event activity likelihood to weight the 
$\{{\intersection}:{\union}\}$ ratio as in \cite{Phan2018b,Phan2017e}, in to order to more aggressively penalize the network.
Finally, the network is trained to minimize the accumulated unweighted multi-task losses over all training examples:
\begin{align}
E= \sum\nolimits_{i}E_{\text{class}}(\mathfrak{G}_i, \hat{\mathfrak{G}}_i) + E_{\text{dist}}(\mathfrak{G}_i, \hat{\mathfrak{G}}_i) + E_{\text{conf}}(\mathfrak{G}_i, \hat{\mathfrak{G}}_i). \label{eq:loss}
\end{align}

\vspace{-0.45cm}
\section{Inference}
\label{sec:inference}
\vspace{-0.2cm}
Inference for joint event detection and boundary estimation can be carried out individually for different event categories of interest similar to \cite{Phan2018a}. However, we need to extend this to handle the sequential output of the network.

Let $m,n >0$ both denote the time indices of a continuous test signal. Given a test audio segment $\mathbf{S}(m)$ of length $T$ frames starting at the time index $m$ and assuming its output sequence $\hat{\mathfrak{G}}(m)$, its contribution to the confidence score that a target event of class $c$ occurs at  time index $n$ is given by
\vspace{-0.3cm}
\begin{align}
f_c\big(n\,|\,\mathbf{S}(m)\big)\!=\!\sum_{t=1}^{T}\hat{y}^c_t(m)\mathds{1}\!\Big(\hat{y}^c_t(m)\!>\!\alpha_c\!\Big)\mathds{1}\!\Big(\!n\!\in\!\Omega\big(p^c_t(m), q^c_t(m)\big)\!\Big),\nonumber
\end{align}

\vspace{-0.2cm}
\noindent where $\alpha_c$ denotes a class-specific threshold on event activity likelihood, $\Omega\big(p^c_t(m), q^c_t(m)\big)$ represents the \emph{region of interest (ROI)} determined by $p^c_t(m)$ and $q^c_t(m)$, and $\mathds{1}(\cdot)$ is the indicator function. 
In essence, we iterate over the output sequence $\hat{\mathfrak{G}}(m)$ and collectively integrate the event activity likelihoods into a confidence score. In addition, the event activity likelihood $\hat{y}^c_t(m)$ at time step $t$ of the sequence is only counted if it is greater than the likelihood threshold $\alpha_c$ and where $n$ lies inside the ROI $\Omega\big(p^c_t(m), q^c_t(m)\big)$, meaning
\begin{align}
m + t - \hat{p}^c_t(m) \le n \le m + t + \hat{p}^c_t(m).
\end{align}
Note that the estimated onset and offset distances need to be denormalized to their original range before inference.

The confidence score obtained from all test audio segments sampled from the test signal is
\vspace{-0.1cm}
\begin{align}
f_c(n) = \sum\nolimits_m f_c\big(n\,|\,\mathbf{S}(m)\big).
\end{align}
 A second class-specific detection threshold $\beta_c$ is applied to the confidence score for joint event detection and segmentation. Although we do not study early detection of an ongoing event~\cite{phan2015b,Phan2018a} in this paper, the inference scheme described has such a capability.

\vspace{-0.2cm}
\section{Experiments}
\vspace{-0.2cm}
\subsection{Datasets}
\vspace{-0.1cm}
We conducted experiments on two datasets: 

{\bf ITC-Irst \cite{Temko07}} \textemdash created for studying isolated AED, this database consists of twelve recording sessions with 16 annotated event categories. Following the standard split used in previous works \cite{Temko07,Phan2015,Phan2018a}, nine out of twelve recordings were used for training and the remaining three were used for evaluation. In addition, evaluation was based on twelve out of 16 categories with the others considered as background sounds. For relevant parameter search (cf. Section \ref{ssec:network_training}), leave-one-recording-out cross-validation was conducted on the nine training recordings. The channel  \emph{TABLE\_1} \cite{Temko07} was chosen for the experiments.

{\bf TUT-SED-Synthetic-2016 \cite{Cakir2017}} \textemdash created for studying overlapping audio event detection, this database consists of 100 mixtures of 994 isolated event instances from 16 event categories. Further detail on the dataset creation procedure can be found in \cite{Cakir2017}. Out of 100 created mixtures, 60 were used for training, 20 for evaluation, and 20 for validation \cite{Cakir2017}.

\vspace{-0.2cm}
\subsection{Network training and parameters}
\label{ssec:network_training}
\vspace{-0.15cm}

To form the training data, we sampled all possible audio segments of length $T$ frames from the training recordings. The network was trained with a minibatch size of four for 100 and 10 epochs for ITC-Irst and TUT-SED-Synthetic-2016, respectively. $F=256$ convolutional filters were used for the convolutional layers and the size of hidden state vectors of the recurrent layer was $H=256$.  The network was trained using the \emph{Adam} optimizer \cite{Kingma2015} with learning rate $10^{-4}$. For regularization, a dropout rate of $0.25$ was applied to the convolutional layers, the recurrent layer, and the residual connection.

Following training, the network was exercised on audio segments sampled from a test signal without overlap to compute the confidence scores as described in Section \ref{sec:inference}. Particularly, for ITC-Irst, we utilized the cross-validation models for this purpose. The final confidence score on the test signal was averaged from the individual ones resulting from the cross-validation models. The detection confidence score was normalized to [0,1] and the category-specific thresholds $\alpha_c$ and $\beta_c$ were selected to maximize the average F1-score on the validation set. $\alpha_c$ and $\beta_c$ were searched in the range [0, 1] with a step size of $0.01$ and $0.05$, respectively.

\vspace{-0.2cm}
\subsection{Evaluation metrics}
\vspace{-0.15cm}
With the proposed multi-label multi-task CRNN coupled with the inference algorithm in Section \ref{sec:inference}, we are interested in detecting entire events and segmenting them from a continuous test signal. Therefore, we evaluated the detection performance based on two event-wise metrics: F1-score and detection error rate (ER).

\vspace{-0.1cm}
\subsection{Baseline}
\vspace{-0.15cm}
In addition to prior works, we implemented a multi-label CRNN baseline for comparison, as it has been demonstrated to achieve state-of-the-art performance on several similar AED datasets \cite{Lim2017,Cakir2017,Kao2018}. The baseline body architecture and parameters were maintained to be the same as the proposed network, except that its output layer only includes multi-label event activity output. As post-processing is important in assisting such a baseline system to yield good performance \cite{Lim2017,Parascandolo2016,Cakir2015}, category-specific likelihood threshold $\alpha_c$ was firstly applied to produce discrete labels which were then smoothed by median filtering. A grid search was conducted for $\alpha_c$ as in Section \ref{ssec:network_training}, while the window size of the median filter was searched in range of [1, 256] with a step of 6. Those values resulting in the best F1-score on the validation set were retained for evaluation.

\begin{figure} [!t]
	\centering
	\includegraphics[width=0.95\linewidth]{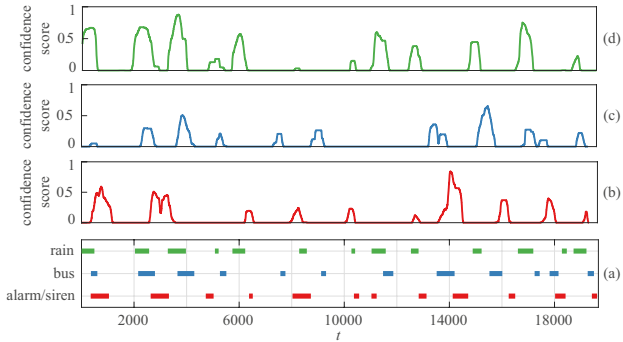}
	\vspace{-0.4cm}
	\caption{The confidence scores for three different categories produced by the proposed network on one test recording of the TUT-SED-Synthetic-2016 dataset. (a) event activity, (b) ``alarms and sirens'', (c) ``bus'', and (d) ``rain''.}
	\label{fig:confidence_score}
	\vspace{-0.2cm}
\end{figure}

\vspace{-0.2cm}
\subsection{Experimental results}
\vspace{-0.1cm}
The results obtained by the proposed multi-label multi-task CRNN and the multi-label CRNN baseline on the two experimental datasets are shown in Tables  \ref{tab:performance_itc} and \ref{tab:performance_tut}. 
To further compare with existing works on the ITC-Irst dataset, we also contrast their performance with the best result previously reported in \cite{Phan2018a} using a dual-DNN approach. 
On the TUT-SED-Synthetic-2016 dataset, no prior event-wise results were reported, so a CRNN-based system, similar to that  
reporting best frame- and segment-wise performance \cite{Cakir2017}, is used for a baseline here.

Results show that the proposed system outperforms the baseline on both isolated and overlapping AED tasks. In the isolated AED case with ITC-Irst, an absolute gain of $0.9\%$ was achieved on both average F1-score (i.e. the event \textit{categories} are considered equally important) and overall F1-score (i.e. the event \textit{instances} are considered equally important) although it brings up the overall ER by $0.8\%$ absolute. 
The rise in ER is mainly due to insertion errors on ``key jingle'' and ``phone ring'', which are likely due to their bi-modal behaviour. The improvement of the proposed system over the baseline becomes even more noticeable on the overlapping data in TUT-SED-Synthetic-2016. Absolute gains of $1.2\%$ on average F1-score and of $4.1\%$ on overall F1-score were achieved. Moreover, it also improves average and overall ER by $2.3\%$ and $18.7\%$ absolute, respectively. 
Fig. \ref{fig:confidence_score} further shows the confidence scores produced by the proposed system for three different event categories on one recording of TUT-SED-Synthetic-2016. Although the events overlap heavily (and events of other categories were also present but are excluded from the plot for clarity), the proposed network seems to be able to untangle the mixtures and recognize the individual instances. 
Meanwhile comparing results in Tables \ref{tab:performance_itc} and \ref{tab:performance_tut}, it is clear that overlapping, or polyphonic, AED remains a more challenging task than isolated AED.
 
On the other hand, the proposed system significantly outperforms the prior work (i.e. the dual-DNN system \cite{Phan2018a}) that reported the best results on ITC-Irst. Absolute gains of $2.5\%$ and $2.1\%$ were achieved on average and overall F1-score, respectively, while average ER was lowered by $1.0\%$ absolute.

\setlength\tabcolsep{2.5pt} 
\begin{table}[!t]
	\caption{ITC-Irst: the detection performances obtained by the proposed approach and the baseline in comparison with the best reported existing work \cite{Phan2018a}.}
	\footnotesize
	\vspace{-0.1cm}
	\begin{center}
	\begin{tabular}{|>{\arraybackslash}m{0.8in}|>{\centering\arraybackslash}m{0.3in}|>{\centering\arraybackslash}m{0.3in}|>{\centering\arraybackslash}m{0.3in}|>{\centering\arraybackslash}m{0.3in}|>{\centering\arraybackslash}m{0.35in}|>{\centering\arraybackslash}m{0.35in}|>{\centering\arraybackslash}m{0in} @{}m{0pt}@{}}
		\cline{1-7}
		\multirow{2}{*}{Event type} & \multicolumn{2}{c|}{Proposed} &  \multicolumn{2}{c|}{Baseline}  & \multicolumn{2}{c|}{\makecell{Best reported \\ (Dual-DNN \cite{Phan2018a})}} & \parbox{0pt}{\rule{0pt}{0ex+\baselineskip}} \\ [0ex]  	
		\cline{2-7}
			
			& F1 & ER & F1 & ER & F1 & ER & \parbox{0pt}{\rule{0pt}{0ex+\baselineskip}} \\ [0ex]  	
			\cline{1-7}
			
			door knock~~~~~& $100.0$ & $0.0$ & $100.0$ & $0.0$ & $100.0$ & $0.0$ &  \parbox{0pt}{\rule{0pt}{\baselineskip}} \\ [0ex]  	
			
			door slam~~~~~~& $100.0$ & $0.0$& $100.0$ & $0.0$    & $100.0$ & $0.0$ & \parbox{0pt}{\rule{0pt}{\baselineskip}} \\ [0ex]  	
			
			steps~~~~~~~~& $100.0$ & $0.0$ & $100.0$ & $0.0$    & $91.7$& $16.7$ &  \parbox{0pt}{\rule{0pt}{\baselineskip}} \\ [0ex]  	
			
			chair moving~~~~&  $92.3$ & $33.3$& $91.7$ & $33.3$   & $92.0$ & $16.7$ &  \parbox{0pt}{\rule{0pt}{\baselineskip}} \\ [0ex]  	
			
			spoon cup jingle~~& $100.0$ & $0.0$& $100.0$ & $0.0$   & $100.0$& $0.0$ &  \parbox{0pt}{\rule{0pt}{\baselineskip}} \\ [0ex]  	
			
			paper wrapping~~& $100.0$ & $0.0$& $100.0$ & $0.0$      & $100.0$&  $0.0$ &  \parbox{0pt}{\rule{0pt}{\baselineskip}} \\ [0ex]  	
			
			key jingle ~~~~~&  $95.7$ & $25.0$& $95.7$ & $8.3$     & $95.7$& $8.3$ &  \parbox{0pt}{\rule{0pt}{\baselineskip}} \\ [0ex]  	
			
			keyboard clicking~&  $96.0$ & $8.3$ & $86.7$ & $33.3$  & $91.7$ & $16.7$ &  \parbox{0pt}{\rule{0pt}{\baselineskip}} \\ [0ex]  	
			
			phone ring~~~~~&   $97.4$ & $30.4$& $98.0$ & $21.7$   & $100.0$& $17.4$ &  \parbox{0pt}{\rule{0pt}{\baselineskip}} \\ [0ex]  	
			
			applause~~~~~& $100.0$ & $0.0$ & $100.0$ & $0.0$  & $100.0$ & $0.0$ & \parbox{0pt}{\rule{0pt}{\baselineskip}} \\ [0ex]  	
			
			cough~~~~~~~&  $93.8$ & $16.7$  & $92.9$ & $16.7$  & $88.0$ &  $25.0$ &  \parbox{0pt}{\rule{0pt}{\baselineskip}} \\ [0ex]  	
			
			laugh~~~~~~~&   $95.7$ & $8.3$ & $95.7$ & $8.3$  & $81.8$ & $33.3$ & \parbox{0pt}{\rule{0pt}{\baselineskip}} \\ [0ex]  	
			
			\cline{1-7}
			{\bf Average~~~}&   $\bm{97.6}$ & $10.2$ & $96.7$ & $10.1$ & $95.1$ & $11.2$ &  \parbox{0pt}{\rule{0pt}{\baselineskip}} \\ [0ex]  	
			{\bf Overall~~~}&   $\bm{97.3}$ & $11.0$ & $96.4$ & $10.2$ & $95.2$ & $11.0$ &  \parbox{0pt}{\rule{0pt}{\baselineskip}} \\ [0ex]  	
			\cline{1-7}
		\end{tabular}
	\end{center}
	\label{tab:performance_itc}
	\vspace{-0.4cm}
\end{table}

\setlength\tabcolsep{2.5pt} 
\begin{table}[!t]
	\caption{TUT-SED-Synthetic-2016: the detection performance obtained by the proposed approach and the baseline system.}
	\footnotesize
	\vspace{-0.1cm}
	\begin{center}
		\begin{tabular}{|>{\arraybackslash}m{0.9in}|>{\centering\arraybackslash}m{0.35in}|>{\centering\arraybackslash}m{0.35in}|>{\centering\arraybackslash}m{0.35in}|>{\centering\arraybackslash}m{0.35in}|>{\centering\arraybackslash}m{0in} @{}m{0pt}@{}}
			\cline{1-5}
			\multirow{2}{*}{Event type} & \multicolumn{2}{c|}{Proposed}  &  \multicolumn{2}{c|}{Baseline} & \parbox{0pt}{\rule{0pt}{0ex+\baselineskip}} \\ [0ex]  	
			\cline{2-5}
			
			& F1 & ER & F1 & ER & \parbox{0pt}{\rule{0pt}{0ex+\baselineskip}} \\ [0ex]  	
			\cline{1-5}
			
			alarms \& sirens & $72.6$ & $50.4$ & $78.7$ & $37.2$ & \parbox{0pt}{\rule{0pt}{\baselineskip}} \\ [0ex]  	
			
			baby crying~~~~& $58.0$ & $97.8$ & $58.9$ & $93.0$  &\parbox{0pt}{\rule{0pt}{\baselineskip}} \\ [0ex]  	
			
			bird singing~~~~&  $63.2$ & $97.5$ & $61.4$ & $89.8$ & \parbox{0pt}{\rule{0pt}{\baselineskip}} \\ [0ex]  	
			
			bus~~~~~~~~~&  $71.1$ & $84.1$ & $62.7$ & $103.7$ & \parbox{0pt}{\rule{0pt}{\baselineskip}} \\ [0ex]  	
			
			cat meowing~~~&  $45.0$ & $130.5$   & $43.8$ & $116.3$ & \parbox{0pt}{\rule{0pt}{\baselineskip}} \\ [0ex]  	
			
			crowd applause~& $51.0$ & $91.9$ & $59.4$ & $91.0$  & \parbox{0pt}{\rule{0pt}{\baselineskip}} \\ [0ex]  	
			
			crowd cheering~ & $71.6$ & $49.5$  & $75.2$ & $43.1$ & \parbox{0pt}{\rule{0pt}{\baselineskip}} \\ [0ex]  	
			
			dog barking~~~& $69.4$ & $72.5$ & $83.4$ & $31.6$ & \parbox{0pt}{\rule{0pt}{\baselineskip}} \\ [0ex]  	
			
			footsteps~~~~~ & $56.4$ & $99.0$   & $46.9$ & $133.6$& \parbox{0pt}{\rule{0pt}{\baselineskip}} \\ [0ex]  	
			
			glass smash~~~ & $60.9$ & $118.6$  & $74.7$ & $59.3$ & \parbox{0pt}{\rule{0pt}{\baselineskip}} \\ [0ex]  	
			
			gun shot~~~~~ & $70.6$ & $72.2$ & $47.7$ & $216.5$  & \parbox{0pt}{\rule{0pt}{\baselineskip}} \\ [0ex]  	
			
			horsewalk~~~~ & $60.2$ & $102.3$ & $49.0$ & $110.0$ & \parbox{0pt}{\rule{0pt}{\baselineskip}} \\ [0ex]  	
			
			mixer~~~~~~~ & $81.0$ & $50.5$  & $86.6$ & $35.3$& \parbox{0pt}{\rule{0pt}{\baselineskip}} \\ [0ex]  	
			
			motorcycle~~~~ & $49.6$ & $89.9$  & $44.3$ & $94.9$& \parbox{0pt}{\rule{0pt}{\baselineskip}} \\ [0ex]  	
			
			rain~~~~~~~~& $69.8$ & $63.4$ & $76.8$ & $42.0$  & \parbox{0pt}{\rule{0pt}{\baselineskip}} \\ [0ex]  	
			
			thunder~~~~~~ & $72.8$ & $77.8$ & $54.8$ & $86.7$ & \parbox{0pt}{\rule{0pt}{\baselineskip}} \\ [0ex]  	
			
			\cline{1-5}
			{\bf Average~~~}& $\bm{64.0}$ & $\bm{84.2}$  & $62.8$ & $86.5$ & \parbox{0pt}{\rule{0pt}{\baselineskip}} \\ [0ex]  	
			{\bf Overall~~~} & $\bm{60.4}$ & $\bm{107.4}$  & $56.3$ & $126.1$ & \parbox{0pt}{\rule{0pt}{\baselineskip}} \\ [0ex]  	
			\cline{1-5}
		\end{tabular}
	\end{center}
	\label{tab:performance_tut}
	\vspace{-0.4cm}
\end{table}

\vspace{-0.2cm}
\section{Conclusion}
\vspace{-0.2cm}
This paper has proposed a multi-label multi-task CRNN in an effort to treat the isolated and overlapping audio event detection tasks homogeneously. 
Built on top of a CRNN architecture, the recurrent output layer of the network was designed to accommodate arbitrary numbers of overlapping sounds, i.e. from isolated to maximally polyphonic (all event categories occurring simultaneously), at every recurrent time step. 
For network training, three sequential losses, including the multi-label classification loss, the distance estimation loss, and the confidence loss, were introduced to penalize the network on both multi-label event activity classification errors and event boundary estimation errors. 
Evaluating on two datasets, namely ITC-Irst for isolated AED and TUT-SED-Synthetic-2016 for overlapping AED, we demonstrated that the proposed network outperforms the multi-class CRNN baseline with the same network body, as well as previously published state-of-the-art results.


\bibliographystyle{IEEEbib}
\bibliography{reference}

\end{document}